\documentclass[10pt,journal,compsoc]{IEEEtran}

\ifCLASSOPTIONcompsoc
  \usepackage[nocompress]{cite}
\else
  \usepackage{cite}
\fi
\ifCLASSINFOpdf
\else
\fi
\usepackage[justification=centering]{caption} 
\usepackage[utf8]{inputenc}
\usepackage{graphicx}
\begin{document}
%

\title{Combining Supervised and Un-supervised Learning for Automatic Citrus Segmentation}
%
%
%
%

\author{Heqing~Huang$^1{^2}$,~
        Tongbin~Huang$^1{^2}$,~
        Zhen~Li$^1{^2}{^3}^*$,~
        Zhiwei~Wei$^1,~$
        and~Shilei~Lv$^1{^2}{^3}$
        
        $^1$ School~of~electronic~engineering(School~of~artificial~intelligence), South~China~Agricultural~University,~China;
        
        $^2$ Mechanization~Research~Office~of~national~citrus~industry~technology~system,~China;
        
        $^3$ Guangdong~Agricultural~Information~Monitoring~Engineering~Technology~ Research~Center,~China

\IEEEcompsocitemizethanks{\IEEEcompsocthanksitem Heqing Huang is with the College of Electronic Engineering, South China Agricultural University, China.\protect\\
E-mail: huangheqing@stu.scau.edu.cn

\IEEEcompsocthanksitem Zhen Li is with the College of Electronic Engineering, South China Agricultural University, China.\protect\\
E-mail: lizhen@scau.edu.cn

}%

}

%
%

\markboth{}%
{Shell \MakeLowercase{\textit{et al.}}: Bare Advanced Demo of IEEEtran.cls for IEEE Computer Society Journals}
%



\IEEEtitleabstractindextext{%
\begin{abstract}

\noindent Citrus segmentation is a key step of automatic citrus picking. While most current image segmentation approaches achieve good segmentation results by pixel-wise segmentation, these supervised learning-based methods require a large amount of annotated data, and do not consider the continuous temporal changes of citrus position in real-world applications. In this paper, we first train a simple CNN with a small number of labelled citrus images in a supervised manner, which can roughly predict the citrus location from each frame. Then, we extend a state-of-the-art unsupervised learning approach to pre-learn the citrus's potential movements between frames from unlabelled citrus's videos. To take advantages of both networks, we employ the multimodal transformer to combine supervised learned static information and unsupervised learned movement information. The experimental results show that combing both network allows the prediction accuracy reached at 88.3$\%$ IOU and 93.6$\%$ precision, outperforming the original supervised baseline 1.2$\%$ and 2.4$\%$. Compared with most of the existing citrus segmentation methods, our method uses a small amount of supervised data and a large number of unsupervised data, while learning the pixel level location information and the temporal information of citrus changes to enhance the segmentation effect.

\end{abstract}

}

\maketitle

\IEEEdisplaynontitleabstractindextext

%
\IEEEpeerreviewmaketitle

\ifCLASSOPTIONcompsoc
\IEEEraisesectionheading{\section{Introduction}\label{sec:introduction}}
\else
\section{Introduction}
\label{sec:introduction}
\fi

\noindent Citrus segmentation \cite{Lee2017An} aims to identify or position citrus from images or videos \cite{2004Citrus,2020Detecting}. This technology is a preliminary step to a wide range of citrus-related applications such as automatic citrus picking \cite{1990Robotic}, fine management and citrus orchards of prediction \cite{2007Prediction}. While most methods detecting citrus based on the single image \cite{XiongJuntao2020,BiSong2019}. Real-world applications usually require to detect citrus from multiple frames (videos) continuously collected, for example, it is necessary to evaluate the high-resolution video frames and calculate the total number of citrus when constructing the citrus yield map \cite{2017A}. Since temporal information can provide objects' dynamics, it is crucial to many video-based recognition/tracking/segmentation applications \cite{2011Track,wang2019learning,dynamic2021self,hu2018unsupervised}. As a result, we also propose to model temporal information for citrus segmentation.

\begin{figure}[ht]
    \centering 
    \includegraphics[width=9cm]{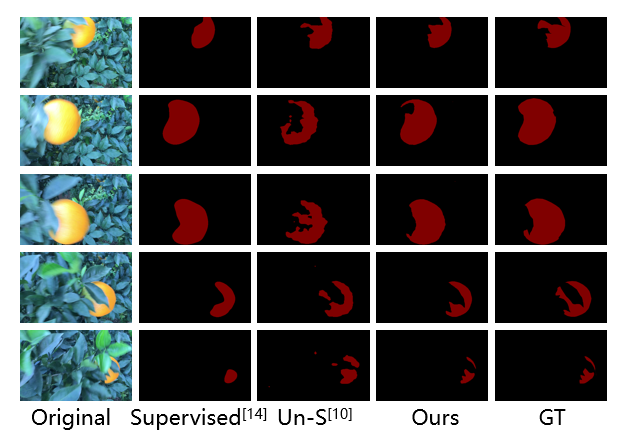}
    \caption{Some examples of our work, from left to right, are: frames extracted from video, results of UNet segmentation, unsupervised temporal modeling of citrus in videos, our approach, manual annotations(GT).}
    \label{fig:galaxy}
\end{figure}

While manually annotating citrus in videos is time-consuming and expensive, training a deep learning model for citrus segmentation usually needs a large number of training samples. Since the movement of citrus in a video is usually smooth and continuous, the locations of any citrus can be very similar in adjacent frames. Thus, we refer to a lot of unsupervised principles \cite{2017Unsupervised,dwibedi2019temporal,2020Space} and design a video detection method. This method takes the time cycle as a round, initialize an image patch randomly, through the forward and backward tracking, the infinite self supervised representation is obtained by matching in deep space. We use this method to learn correspondence and temporal information from a lot of unsupervised data. Then, we trained the supervised semantic segmentation model \cite{2015U,2017Pyramid,2015Fully,2018Encoder} to get the prediction chart. Finally, we can potentially learn the elements between different prediction results through attention. Specifically, we use point-to-point pixel segmentation and the correspondence learned in the video to fuse them through a converter formed by attention mechanism. In other words, this method only uses a small amount of labeled data and does not need expensive labeling cost. It combines supervised learning with unsupervised learning, and uses attention mechanism to get better segmentation results.

To evaluate the proposed approach, we specifically collected a citrus segmentation dataset and make it publicly available, which is another contribution of this paper. The dataset contains $1513$ annotated citrus images and a large number of unlabelled citrus videos. we have compared our approach to existing supervised and unsupervised approaches. On this dataset, our approach generated the IOU result of 88.3$\%$ and the pixel accuracy of 93.6$\%$ by combining the unsupervised method with the supervised method based on UNet \cite{2015U}, as well as the IOU result of 85.3$\%$ and the pixel accuracy of 92.1$\%$ by combining the unsupervised method based on the PSPNet \cite{2017Pyramid}, which are clearly outperformed the results achieved by supervised baselines. Our segmentation example is shown in Figure 1.

The main contributions of our study can be summarized as follows: 1) We propose a weakly supervised learning method to combine the point-to-point pixel segmentation and temporal continuity provided by videos, it not only provides rich time information, but solves the problem of low utilization of a large number of unlabeled video data in real scenes; 2) We provide 1513 citrus images with annotation and 300 citrus videos dataset.

\section{Related work}
This section presents a comprehensive review on computer vision-based methods for conventional citrus object detection and its development in deep learning (from sect. \ref{sec:method} to sect. \ref{sec:deep}). This section gives a brief introduction to the methods used in this paper, like video detection, self-supervised learning.(sect. \ref{sec:video} to sect. \ref{sec:unsupervised}).

\subsection{Traditional method}\label{sec:method}
Early citrus recognition studies under natural conditions were mainly focused on traditional image processing algorithms. For example, Investigation by Hissin et al \cite{2012Digital} showed that a effective method of circular Hough transform for target detection. Dorj et al \cite{2017An} proposed different color features in images to predict the yield of citrus that converted RGB images to HSV images, then distinguish the color of citrus. These methods have high complexity and there are difficult to accurately identify the citrus under the condition of dense accumulation and leaves occlusion.

\subsection{Deep learning method}\label{sec:deep}
Many existing deep learning methods also achieve good segmentation results in the field of citrus detection. Deng et al \cite{dung2020} use Mask R-CNN \cite{2017Mask} and optimizes the main convolution part and mask branch part to achieve efficient detection of dense small-scale citrus flowers in complex structure images. Bi et al \cite{BiSong2019} used multiple segmentation method to improve the multi-scale image detection ability and real-time performance of citrus target recognition model, then complete the training of the citrus target recognition model based on transfer learning. Xiong et al \cite{XiongJuntao2020} referred to the residual network and dense connection network, which realizes the reuse and fusion of multi-layer features of the network, and enhances the robustness of small target and overlapping occlusion fruit recognition, with an accuracy rate of 97.67$\%$. Hu et al \cite{Hu2019} proposed a method for segmentation and recognition of ripe citrus fruits with regional characteristics by using feature mapping to reduce the dimension based on color characteristics of images.\par

\subsection{Video object segmentation}\label{sec:video}
Video detection use temporal context information to enhance performance, focusing on temporal context information can eliminate problems such as motion blurring or occlusion and small object areas in some single frame images. On the other hand, the optical flow between two frames is an ideal feature for video, including the motion track of the target object, and it has also achieved good accuracy in video detection. With the idea of attention mechanism, Flow Guided Feature Aggregation \cite{2017} calculates the cosine similarity of the difference between the current frame and following frames as adaptive weight, and learns the network model, which improves the feature quality and improves problem of motion blur in videos. Gan et al \cite{2018An} used the video captured by a thermal camera to capture the surface temperature of citrus canopy to detect whether citrus fruit are ripe or not, which provides a more novel method for fruit detection and tracking. Liu et al \cite{2020FCRN} combined with depth subdivision, frames-to-frames tracking and 3D localization to accurately count the fruits in image sequence, used the full convolution network to segment the fruit pixel in the video frame, and tracked the fruit locus, which has high accuracy and robustness.\par

\subsection{Self-supervised learning}\label{sec:unsupervised}
\noindent Since natural objects' movements are usually smooth and continuous, they are underlying relationships of the objects' locations and gestures in adjacent frames. As a result, many previous studies \cite{dwibedi2019temporal,2017Unsupervised,wang2019learning,2020Space,liu2019selflow,song2019inferring,pathak2017learning} have investigated to learning objects/human motions in un-supervised/self-supervised manners. Dwibedi et al \cite{dwibedi2019temporal} designed a TCC network that provides a self-monitoring video detection model. Learning representation through the connection between parallel processes example, which can be applied to understand the fine-definition features of video. According to the motion dependence in video, Luo et al \cite{2017Unsupervised} designed encoder-decoder network based on recursive neural network to predict the atomic 3D stream sequence of RGB-D modal computation, which reduces the complexity of learning framework and can capture the robust video features of long-term motion dependence and spatio-temporal relationship. Wang et al \cite{wang2019learning} designed a temporal consistency to serve as a freely supervised signal to maintain the consistency of recognition by forward tracking and backward tracking, and to be competitive with strongly supervised learning, which can encode objects' dynamics. Ajabri et al \cite{2020Space} treated multiple frames in a video as a sequence of spatio-temporal graphs, and shares directional edges in temporally adjacent nodes to learn representational features by random wandering and temporal consistency in the absence of data labels in the video.\par


\begin{figure*}[ht]
    \centering 
    \includegraphics[width=18cm]{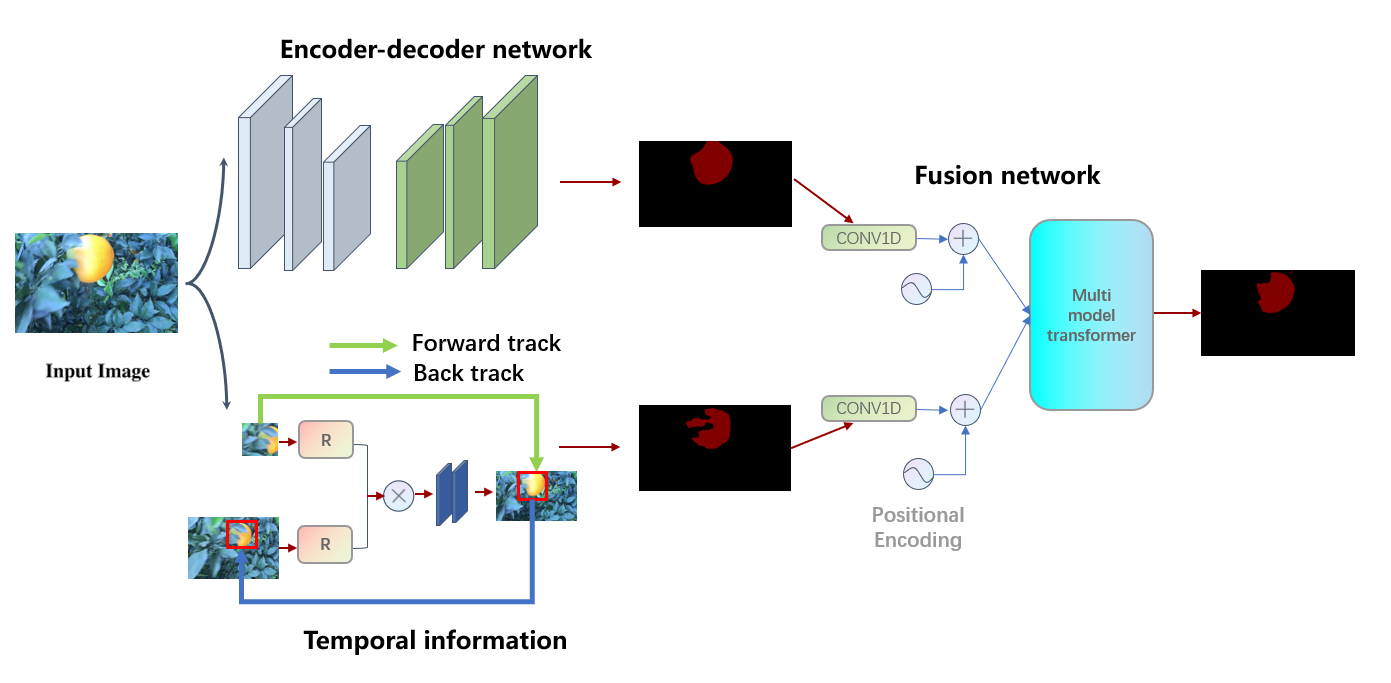}
    \caption{Our model is divided into three main parts: supervised learning, unsupervised learning and fusion network using static and temporal features.}
    \label{fig:galaxy}
\end{figure*}

\section{The proposed method}

\noindent The proposed approach consists of three modules: an encoder-decoder network trained with annotated static citrus images (sect. \ref{sec:su}), a temporal network that learns temporal correspondences between frames, which provides temporal constraints of the detected citrus between frames (Sect. \ref{sec:unsu}) and a fusion network that takes advantages of both static frame-based citrus prediction and citrus's temporal correspondence between frames (Sect. \ref{sec:muti}). Consequently, the learnt temporal network can provide extra temporal correspondence information helping the supervised learned network to better segment citrus. Importantly, this temporal network is trained in an unsupervised manner without requiring any labels.

\subsection{Supervised learning of citrus segmentation network}\label{sec:su}
\noindent In this section, we trains four popular generative network for the supervised learning-based citrus segmentation. In particular, we employ FCN \cite{2015Fully}, UNet \cite{2015U}, PSPNet \cite{2017Pyramid}, DeepLabv3 \cite{2018Encoder}, as they have been widely applied to various image generation applications and achieved excellent performance. Among these networks, Unet has achieved the best results, it can combine high-level and low-level features to provide context semantic information in the process of segmentation, the semantics is simple and clear, and the model parameters are few, which is beneficial to the segmentation of citrus.

We annotate a small number of citrus images, use label to annotate, extract each citrus contour, fine tune them based on the pre-training weight, use the structure of encoder-decoder with ResNet-50 \cite{he2016deep} as the feature extractor. We set the loss function as BCE with logits loss, use this model to roughly determine the position of each frame of citrus video.

\subsection{Unsupervised temporal modeling of citrus in videos}\label{sec:unsu}

\noindent In the practical application of citrus picking and positioning, the position of the detected citrus, the rotation degree and light conditions of the citrus can be changed between frames. However, the CNNs that trained with annotated static images can not capture such temporal information which are crucial to position and segment citrus. While manually annotate video frames is time-consuming and expensive, we propose to learn such information from a large number of unlabeled videos in a unsupervised manner.

Our model uses a self-monitoring method to learn the correspondence of virtual language, and uses the time cycle consistency as the free monitoring signal to complete the target segmentation. We cut 80 * 80 image blocks randomly from 256 * 256 video frames, use Resnet50 to extract features, take middle-level features of time context in video for block matching, calculate affinity matrix by L2 normalization and point multiplication, then add convolution layer for affine transformation, and self supervise learning embedded features of each frame. The model by tracking backward and then forward, and takes the inconsistency between the starting point and the ending point as the loss function. In order to minimize the loss and maintain the consistency of recognition, through cross frame recognition to learn and optimization, get the correspondence of citrus changes in practical application.

\subsection{ Citrus segmentation using static and temporal features}\label{sec:muti}
\noindent In order to segment and locate citrus more accurately, we propose a weakly supervised learning method, which uses the temporal constraints between consecutive frames provided by the unsupervised model to constrain the citrus prediction images obtained by the supervised learned encoder-decoder network. We uses converter to learn the relationship between the two prediction images, so as to focus on the concerns of citrus images, it makes the segmentation result more accurate.
 
Considering the different data distribution between different results, the cross model converter is used to construct the interaction information between different models. Through one-dimensional convolution dimension reduction, the two input features are mapped to the same dimension, and segmentation predictions of supervised learned model can be further enhanced by the such temporal constraints. After obtaining the temporal information, we add the location information to encode it, use the transformer structure based on self attention to model, and map and predict the structural features of the spliced transformer. The fusion network is optimized and iterated by supervised learning of prediction results and labels.

\begin{figure*}[ht]
    \centering 
    \includegraphics[width=18.3cm]{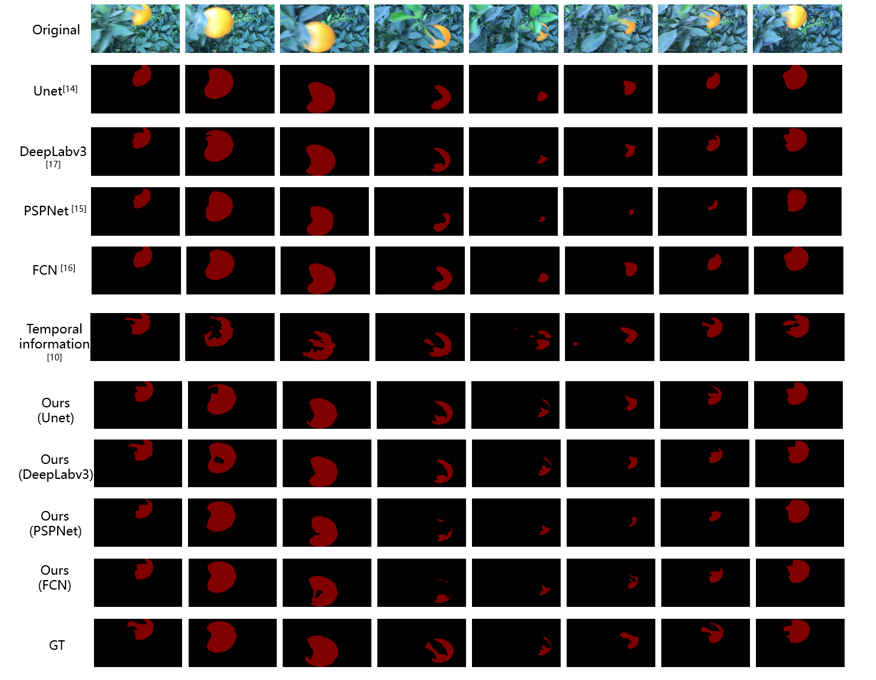}
    \label{fig:galaxy}
    \caption{We show the visualization effect of the following methods and manual annotation: four supervised learning models of citrus segmentation network, an unsupervised learning model to obtain temporal information from videos, and the model combines static and temporal information.}
    \label{fig:galaxy}
\end{figure*}

\section{Experiments}
In this section, we will describe the different types of data we collect. We show the training details of different models and report the performance of each model. Then summarize the ablations for other components of our method. For the sake of unified standards, the three model test sets we use are all the test sets of unsupervised models.

\subsection{Databases}
\noindent We evaluated supervised method on the citrus image dataset. The image database consists of 1513 citrus images and labels with a resolution of 1920*1080, it was collected at a distance of one meter from citrus fruit trees under natural light. Considering the corresponding relationship between labels and data and ensuring uniform distribution of database, we use 1200 citrus images as training set and 313 citrus images as verification set.

In order to get temporal information, we collected dataset containing citrus videos. The video database consists of recorded videos in which the citrus being picked and shaken. In normal light, we collected 300 videos, with an average of 30-40 seconds per video, the frame rate is 30 frames per second. We selected 60 videos, extracted one image every 5 frame, labeled them manually. And set 20 videos as the test set. We use the remaining 240 videos as the training data of the unsupervised model training. 

For fusion networks, we use the prediction results generated by supervised model and unsupervised model, and take 30 video results as the training set and 10 videos as the verification set.

\begin{figure*}[htp]
    \centering 
    \includegraphics[width=17cm]{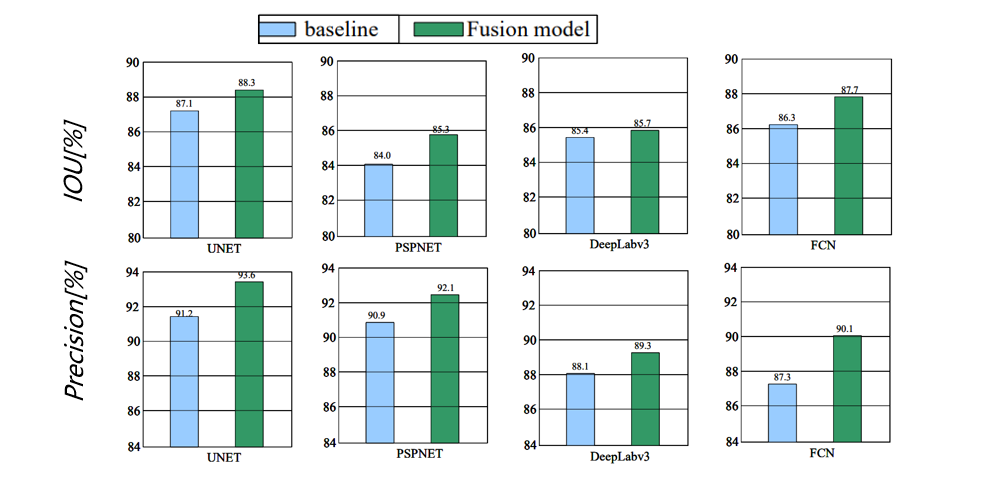}
    \centering \caption{We tested four supervised learning segmentation methods and our methods.}
    \centering
    \label{fig:galaxy}
\end{figure*}

\subsection{Implementation detail and metrics}
\noindent Training. The training environment of all models is Ubuntu 18.04 and the processor is e5-2620 V4@2.10GHz, 8-core, 16GB ram, NVIDIA geforce RTX 2080ti. We train the following four supervised models (FCN\cite{2015Fully}, UNet\cite{2015U}, DeepLabv3\cite{2018Encoder}, PSPNet\cite{2017Pyramid}) with batch size of 8, for 60 epochs. For unsupervised model, We first extract the video frames and adjust the resolution of each frame to 256 * 256. We use the basic feature to extract network resnet50, set the optimizer as SGD and the learning rate as 0.0002. When training the fusion model, we set batch size is 4, the learning rate as 0.003 and the optimizer is Adam.

Inference. At test time, we use the unsupervised network and give the initial label of the first frame to pass it into other frames of the video. At the same time, we make the trained supervised model get the segmentation results of the test set and put them into the fusion network for testing. We follow the standard metrics including precision (P) and intersection over union (IOU). P is defined as follows:
\begin{equation}
    P = \frac{num\_pixel(pred = gt)}{num\_pixel(total)}
\end{equation}
While $num\_pixel(pred = gt)$ means predicted value equals the number of pixels in the real label, $num\_pixel(total)$ is the number of pixels in the image, $pred$ and $true$ represent the predicted pixel value and the real label of citrus.
IOU is be defined as:
\begin{equation}
    IOU = \frac{Area\quad of \quad Overlap}{Area \quad of\quad  Union} = \frac{A_{pred} \cap A_{true}}   {A_{pred} \cup A_{true}}
\end{equation}
While $A_{pred}$ means the predicted citrus region, $A_{true}$ means the real citrus region.

\subsection{Comparison to existing approaches}
\noindent Table 1. compares the proposed combination approach with several four supervised baselines and a unsupervised baseline. Firstly, it is clear that both supervised and unsupervised approaches can already provide good segmentation performance, where supervised models still clearly better than the unsupervised model. In addition, despite supervised trained UNet and PSPNet already achieved over $90\%$ precision, the proposed approach uses the unsupervised-learned citrus movement information to bring extra benefit, allowing it achieving the best performance over all evaluated models, which demonstrated the capability of the proposed approach.

\begin{table}[h]
    \centering
    {
    \begin{tabular}{|c|c|c|}
    \hline 
    supervised models & P & IOU\\
    \hline
    \ FCN\cite{2015Fully} & 0.873 & 0.863 \\
    \ PSPNet\cite{2017Pyramid} & 0.909 & 0.840 \\
    \ DeepLabv3\cite{2018Encoder} & 0.879 & 0.854 \\
    \ UNet\cite{2015U} & 0.912 & 0.871 \\
    \ Unsupervised model\cite{wang2019learning} & 0.789 & 0.704 \\
    \ Ours & 0.936 & 0.883 \\
    \hline 
    \end{tabular}
    }
    \caption{We tested four supervised learning models and unsupervised learning citrus segmentation results under the same standard.}
    \label{tab:my_label}
\end{table}

We use a video based unsupervised model to learn temporal information and correspondence, and provide test indicators, detailed in \cite{2016A}, results as seen in Table 2. 

\begin{table}[h]
    \centering
    {
    \begin{tabular}{|c|c|c|c|c|c|}
    \hline
         J-mean	& J-recall & J-decay & F-mean & F-recall & F-decay\\
         \hline
         0.704 & 0.898 & 0.062& 0.663 & 0.779 & 0.120\\
    \hline
    \end{tabular}
    }
    \caption{We verified the effect of our citrus testset on the standard of video segmentation.}
    \label{tab:my_label}
\end{table}

\subsection{Ablation studies}
\noindent In order to show the contribution from fusion model, we design four segmentation baselines to compare the results of our fusion network. In each fusion network, it is better than the baseline based method in the citrus test set. For example, compared with UNET, the fusion networks get P and IOU can reach 93.6$\%$ and 88.3$\%$ which are 2.4$\%$ and 1.2$\%$ higher than the original model. For FCN, our model a 2.8$\%$ increase in precision and 1.4$\%$ in IOU. Similarly, we do the same operation on DeepLabv3, which achieves the precision of 89.3$\%$ and IOU of 85.7$\%$. We also tested our model on PSPNet, the indicators were improved by 1.2$\%$ point and 1.3$\%$ point, as shown in Figure 4, where the green histogram is our method, and the blue is the original baselines. These results show that our fusion model can get better segmentation effect than baseline.

\noindent In order to facilitate comparison, we use a simple weighted fusion method to stack the results of the supervised model and unsupervised model, then compare them with fusion model. We use the UNet model to compare it with the results shown in Table 3. We think that the common fusion method can not learn the characteristics of the two results, which proves the feasibility of using the fusion model again. We visualize our model in Figure 3.

\begin{table}[h]
    \centering
    {
    \begin{tabular}{|c|c|c|}
    \hline 
    supervised models & P & IOU\\
    \hline
    \ Weighted mean & 0.814 & 0.769 \\
    \ Ours & 0.936 & 0.883 \\
    \hline 
    \end{tabular}
    }
    \caption{We used a simple image weighted average to compare our model.}
    \label{tab:my_label}
\end{table}

\section{Conclusion}
\noindent We propose a cooperative citrus segmentation method for video of learning temporal information and pixel level generation at the same time. The architecture consists of three parts: (1) supervised baselines: we use several encoder-decoder baselines to roughly detect the position of citrus in each frame. (2) unsupervised model: we design an unsupervised video segmentation method to learn the change rule of Citrus between adjacent frames, and take it as a constraint. (3) Fusion model: we use the fusion network with attention mechanism to combine the correspondence and location information of citrus, improve the segmentation results. We evaluated the proposed method on the collected image/video citrus dataset, showing that it is better than all supervised learned baselines.

As explained above, the performance of add pixel position and temporal information is available, we try our best to reduce the trouble of manual annotation and achieve good results with a small amount of annotation. We hope that the method in this work can be helpful to the field of citrus segmentation.


%


\ifCLASSOPTIONcaptionsoff
  \newpage
\fi

\bibliographystyle{ieeetr} 
\bibliography{reference} 

\begin{thebibliography}{10}

\bibitem{Lee2017An}
Lee, Malrey, Yun, Sang-seok, Dorj, and Ulzii-Orshikh, ``An yield estimation in
  citrus orchards via fruit detection and counting using image processing,''
  {\em Computers and Electronics in Agriculture}, 2017.

\bibitem{2004Citrus}
H.~Xu and Y.~Ying, ``Citrus fruit recognition using color image analysis,''
  {\em Proceedings of Spie the International Society for Optical Engineering},
  vol.~5608, pp.~321--328, 2004.

\bibitem{2020Detecting}
W.~Chen, S.~Lu, B.~Liu, G.~Li, and T.~Qian, ``Detecting citrus in orchard
  environment by using improved yolov4,'' {\em Scientific Programming},
  vol.~2020, no.~1, pp.~1--13, 2020.

\bibitem{1990Robotic}
R.~C. Harrell, P.~D. Adsit, R.~D. Munilla, and D.~C. Slaughter, ``Robotic
  picking of citrus,'' {\em Robotica}, vol.~8, no.~04, pp.~269--278, 1990.

\bibitem{2007Prediction}
X.~Ye, K.~Sakai, M.~Manago, S.~I. Asada, and A.~Sasao, ``Prediction of citrus
  yield from airborne hyperspectral imagery,'' {\em Precision Agriculture},
  vol.~8, no.~3, pp.~111--125, 2007.

\bibitem{XiongJuntao2020}
X.~Juntao, Z.~Zhenhui, L.~Jiaen, Z.~Zhuo, L.~Bolin, and sun Baoxia, ``orange
  recognition method in night environment based on improved yolo v3 network,''
  {\em Journal of agricultural machinery}, vol.~v.51, no.~04, pp.~206--213,
  2020.

\bibitem{BiSong2019}
B.~song, G.~Feng, C.~Junwen, and Z.~Lu, ``citrus target recognition method
  based on deep convolution neural network,'' {\em Journal of agricultural
  machinery}, vol.~50, no.~05, pp.~188--193, 2019.

\bibitem{2017A}
H.~Gan, W.~S. Lee, and V.~Alchanatis, ``A prototype of an immature citrus fruit
  yield mapping system,'' 2017.

\bibitem{2011Track}
J.~Lezama, K.~Alahari, J.~Sivic, and I.~Laptev, ``Track to the future:
  Spatio-temporal video segmentation with long-range motion cues,'' in {\em
  Computer Vision \& Pattern Recognition}, 2011.

\bibitem{wang2019learning}
X.~Wang, A.~Jabri, and A.~A. Efros, ``Learning correspondence from the
  cycle-consistency of time,'' in {\em Proceedings of the IEEE Conference on
  Computer Vision and Pattern Recognition}, pp.~2566--2576, 2019.

\bibitem{dynamic2021self}
S.~Song, S.~Jaiswal, E.~Sanchez, G.~Tzimiropoulos, L.~Shen, and M.~Valstar,
  ``Self-supervised learning of person-specific facial dynamics for automatic
  personality recognition,'' {\em IEEE Transactions on Affective Computing},
  2021.

\bibitem{hu2018unsupervised}
Y.-T. Hu, J.-B. Huang, and A.~G. Schwing, ``Unsupervised video object
  segmentation using motion saliency-guided spatio-temporal propagation,'' in
  {\em Proceedings of the European conference on computer vision (ECCV)},
  pp.~786--802, 2018.

\bibitem{2017Unsupervised}
Z.~Luo, B.~Peng, D.~A. Huang, A.~Alahi, and L.~Fei-Fei, ``Unsupervised learning
  of long-term motion dynamics for videos,'' in {\em 2017 IEEE Conference on
  Computer Vision and Pattern Recognition (CVPR)}, 2017.

\bibitem{dwibedi2019temporal}
D.~Dwibedi, Y.~Aytar, J.~Tompson, P.~Sermanet, and A.~Zisserman, ``Temporal
  cycle-consistency learning,'' in {\em Proceedings of the IEEE Conference on
  Computer Vision and Pattern Recognition}, pp.~1801--1810, 2019.

\bibitem{2020Space}
A.~Jabri, A.~Owens, and A.~A. Efros, ``Space-time correspondence as a
  contrastive random walk,'' 2020.

\bibitem{2015U}
O.~Ronneberger, P.~Fischer, and T.~Brox, ``U-net: Convolutional networks for
  biomedical image segmentation,'' 2015.

\bibitem{2017Pyramid}
H.~Zhao, J.~Shi, X.~Qi, X.~Wang, and J.~Jia, ``Pyramid scene parsing network,''
  in {\em 2017 IEEE Conference on Computer Vision and Pattern Recognition
  (CVPR)}, 2017.

\bibitem{2015Fully}
J.~Long, E.~Shelhamer, and T.~Darrell, ``Fully convolutional networks for
  semantic segmentation,'' {\em IEEE Transactions on Pattern Analysis and
  Machine Intelligence}, vol.~39, no.~4, pp.~640--651, 2015.

\bibitem{2018Encoder}
L.~C. Chen, Y.~Zhu, G.~Papandreou, F.~Schroff, and H.~Adam, ``Encoder-decoder
  with atrous separable convolution for semantic image segmentation,'' {\em
  Springer, Cham}, 2018.

\bibitem{2012Digital}
R.~Hussin, M.~R. Juhari, N.~W. Kang, R.~C. Ismail, and A.~Kamarudin, ``Digital
  image processing techniques for object detection from complex background
  image,'' {\em Procedia Engineering}, vol.~41, no.~none, pp.~340--344, 2012.

\bibitem{2017An}
U.~O. Dorj, M.~Lee, and S.~S. Yun, ``An yield estimation in citrus orchards via
  fruit detection and counting using image processing,'' {\em Computers and
  Electronics in Agriculture}, vol.~140, pp.~103--112, 2017.

\bibitem{dung2020}
D.~Ying, W.~Huarui, and Z.~Huaji, ``citrus flower recognition and flower count
  statistics based on instance segmentation,'' {\em Journal of Agricultural
  Engineering}, vol.~036, no.~007, pp.~200--207, 2020.

\bibitem{2017Mask}
K.~He, G.~Gkioxari, P.~Dollár, and R.~Girshick, ``Mask r-cnn,'' in {\em 2017
  IEEE International Conference on Computer Vision (ICCV)}, 2017.

\bibitem{Hu2019}
H.~Youcheng, X.~Hongbin, H.~Lin, L.~SA, and Y.~Changhui, ``segmentation and
  recognition of mature citrus and branches and leaves based on regional
  characteristics(in chinese),'' {\em modern manufacturing engineering}, no.~5,
  p.~13, 2019.

\bibitem{2017}
X.~Zhu, Y.~Wang, J.~Dai, L.~Yuan, and Y.~Wei, ``[ieee 2017 ieee international
  conference on computer vision (iccv) - venice (2017.10.22-2017.10.29)] 2017
  ieee international conference on computer vision (iccv) - flow-guided feature
  aggregation for video object detection,'' pp.~408--417, 2017.

\bibitem{2018An}
H.~Gan, W.~S. Lee, V.~Alchanatis, and A.~Abd-Elrahman, ``An active thermography
  method for immature citrus fruit detection,'' in {\em 14th International
  Conference on Precision Agriculture}, 2018.

\bibitem{2020FCRN}
L.~E. C.~L. Rosa, M.~Zortea, B.~H. Gemignani, D.~A.~B. Oliveira, and R.~Q.
  Feitosa, ``Fcrn-based multi-task learning for automatic citrus tree detection
  from uav images,'' 2020.

\bibitem{liu2019selflow}
P.~Liu, M.~Lyu, I.~King, and J.~Xu, ``Selflow: Self-supervised learning of
  optical flow,'' in {\em Proceedings of the IEEE/CVF Conference on Computer
  Vision and Pattern Recognition}, pp.~4571--4580, 2019.

\bibitem{song2019inferring}
S.~Song, E.~S{\'a}nchez-Lozano, L.~Shen, A.~Johnston, and M.~Valstar,
  ``Inferring dynamic representations of facial actions from a still image,''
  {\em arXiv preprint arXiv:1904.02382}, 2019.

\bibitem{pathak2017learning}
D.~Pathak, R.~Girshick, P.~Doll{\'a}r, T.~Darrell, and B.~Hariharan, ``Learning
  features by watching objects move,'' in {\em Proceedings of the IEEE
  Conference on Computer Vision and Pattern Recognition}, pp.~2701--2710, 2017.

\bibitem{he2016deep}
K.~He, X.~Zhang, S.~Ren, and J.~Sun, ``Deep residual learning for image
  recognition,'' in {\em Proceedings of the IEEE conference on computer vision
  and pattern recognition}, pp.~770--778, 2016.

\bibitem{2016A}
F.~Perazzi, J.~Pont-Tuset, B.~Mcwilliams, L.~V. Gool, and A.~Sorkine-Hornung,
  ``A benchmark dataset and evaluation methodology for video object
  segmentation,'' in {\em 2016 IEEE Conference on Computer Vision and Pattern
  Recognition (CVPR)}, 2016.

\end{thebibliography}
\end{document}